%% file: main.tex
\documentclass[letterpaper]{article} 
\usepackage{aaai2026}
\usepackage{times}  
\usepackage{helvet}  
\usepackage{courier}  
\usepackage[hyphens]{url}  
\usepackage{graphicx} 
\usepackage{natbib}  
\usepackage{caption} 
\usepackage{algorithm}
\usepackage{algorithmic}
\usepackage{amsmath}
\usepackage{bm}
\usepackage{booktabs}
\usepackage{multirow}
\usepackage{pifont}
\usepackage{adjustbox}
\usepackage{amssymb}
\usepackage{xcolor}
\usepackage{newfloat}
\usepackage{listings}
\DeclareCaptionStyle{ruled}{labelfont=normalfont,labelsep=colon,strut=off} 
\floatstyle{ruled}
\newfloat{listing}{tb}{lst}{}
\floatname{listing}{Listing}
\title{QuEPT: Quantized Elastic Precision Transformers with One-Shot Calibration for Multi-Bit Switching}
\author{
    Ke Xu\textsuperscript{\rm 1,2}, 
    Yixin Wang\textsuperscript{\rm 2},  
    Zhongcheng Li\textsuperscript{\rm 4}, 
    Hao Cui\textsuperscript{\rm 4}, 
    Jinshui Hu\textsuperscript{\rm 4},  
    Xingyi Zhang*\textsuperscript{\rm 1,\rm 3} \\
}
\affiliations{
    \textsuperscript{\rm 1}State Key Laboratory of Opto-Electronic Information Acquisition and Protection Technology, Anhui University, Hefei, China\\
    \textsuperscript{\rm 2}School of Artificial Intelligence, Anhui University, Hefei, China\\
    \textsuperscript{\rm 3}School of Computer Science and Technology, Anhui University, Hefei, China \\
    \textsuperscript{\rm 4}iFLYTEK Research, Hefei, China \\
    {xuke}@ahu.edu.cn, yxwang@stu.ahu.edu.cn, \{zcli19, haocui, jshu\}@iflytek.com, xyzhanghust@gmail.com
}

\usepackage{bibentry}

\begin{document}

\maketitle
\input{sections/0_abstract}
\input{sections/1_introduction}

\input{sections/2_related_works}

\input{sections/3_method}
\input{sections/4_experiments}

\input{sections/5_conclusion}

\section{Acknowledgments} 
This work was supported in part by the National Natural Science Foundation of China (No. 62576001, No. 62206003, No. U21A20512) and the National Key Research and Development Project (NO. 2018AAA0100105).

\bibliography{main}

\setlength{\leftmargini}{20pt}
\makeatletter\def\@listi{\leftmargin\leftmargini \topsep .5em \parsep .5em \itemsep .5em}
\def\@listii{\leftmargin\leftmarginii \labelwidth\leftmarginii \advance\labelwidth-\labelsep \topsep .4em \parsep .4em \itemsep .4em}
\def\@listiii{\leftmargin\leftmarginiii \labelwidth\leftmarginiii \advance\labelwidth-\labelsep \topsep .4em \parsep .4em \itemsep .4em}\makeatother

\setcounter{secnumdepth}{0}
\renewcommand\thesubsection{\arabic{subsection}}
\renewcommand\labelenumi{\thesubsection.\arabic{enumi}}

\newcounter{checksubsection}
\newcounter{checkitem}[checksubsection]

\newcommand{\checksubsection}[1]{%
  \refstepcounter{checksubsection}%
  \paragraph{\arabic{checksubsection}. #1}%
  \setcounter{checkitem}{0}%
}

\newcommand{\checkitem}{%
  \refstepcounter{checkitem}%
  \item[\arabic{checksubsection}.\arabic{checkitem}.]%
}
\newcommand{\question}[2]{\normalcolor\checkitem #1 #2 \color{blue}}
\newcommand{\ifyespoints}[1]{\makebox[0pt][l]{\hspace{-15pt}\normalcolor #1}}

\end{document}

%% file: sections/0_abstract.tex
\begin{abstract}
Elastic precision quantization enables multi-bit deployment via a single optimization pass, fitting diverse quantization scenarios.
Yet, the high storage and optimization costs associated with the Transformer architecture, research on elastic quantization remains limited, particularly for large language models.
This paper proposes QuEPT, an efficient post-training scheme that reconstructs block-wise multi-bit errors with one-shot calibration on a small data slice. It can dynamically adapt to various predefined bit-widths by cascading different low-rank adapters, and supports real-time switching between uniform quantization and mixed precision quantization without repeated optimization.
To enhance accuracy and robustness, we introduce Multi-Bit Token Merging (MB-ToMe) to dynamically fuse token features across different bit-widths, improving robustness during bit-width switching. Additionally, we propose Multi-Bit Cascaded Low-Rank adapters (MB-CLoRA) to strengthen correlations between bit-width groups, further improve the overall performance of QuEPT.
Extensive experiments demonstrate that QuEPT achieves comparable or better performance to existing state-of-the-art post-training quantization methods. 
\end{abstract}

\begin{links} 
\link{Code}{https://github.com/xuke225/QuEPT}
\end{links}

%% file: sections/1_introduction.tex
\section{Introduction}
Model quantization reduces computational and memory requirements by representing weights and activations using low-bit fixed-point numbers, enabling deep neural networks to run efficiently on resource-constrained devices. However, most quantization methods~\cite{adaround,pdquant,awq,flatquant,mbq} are only able to achieve a single predefined bit-width for quantization, and modifying the bit-width necessitates re-optimization. Elastic precision quantization can jointly optimize multiple predefined bit widths while sharing the original weights. This approach enables a quantized model to dynamically adapt to various bit-widths through one-shot calibration, offering flexible precision-efficiency trade-offs while supporting real-time switching between uniform and mixed-precision configurations without requiring separate optimizations for each bit setting.

While multi-bit quantization has yielded impressive gains for Convolutional Neural Networks (CNNs)~\cite{robustquant,any-precision,eqnet, ptmq,truncquant}, its benefits do not readily carry over to Transformers. Compared with CNNs, Transformers exhibit dense inter-token dependencies, attention-driven dynamic sparsity and expansive activation ranges, all of which magnify quantization noise and make per-layer bit-width sensitivity highly heterogeneous. Consequently, existing efforts to adapt multi-bit techniques to Transformers remain scarce, with most studies still anchored to fixed-bit-width schemes. 
Only very recently have a handful of works begun to explore multi-bit quantization for Transformer-based LLMs, including Any-Precision LLM~\cite{any-precision-llm}, which leverages truncated bit-widths and incremental upscaling, and MatQuant~\cite{matquant}, which employs co-training and co-distillation regularization. However, these methods still underperform state-of-the-art single-bit-width quantization methods~\cite{quarot,spinquant,flatquant} at low precision. Any-Precision LLM only implements weight quantization and does not consider activation quantization. Current approaches also fail to address how aggressive low-bit quantization negatively impacts medium and high-bit representations, resulting in overall performance being constrained by the lowest precision configuration.

\begin{figure*}[!htb]
\centering
\includegraphics[width=0.965\linewidth]{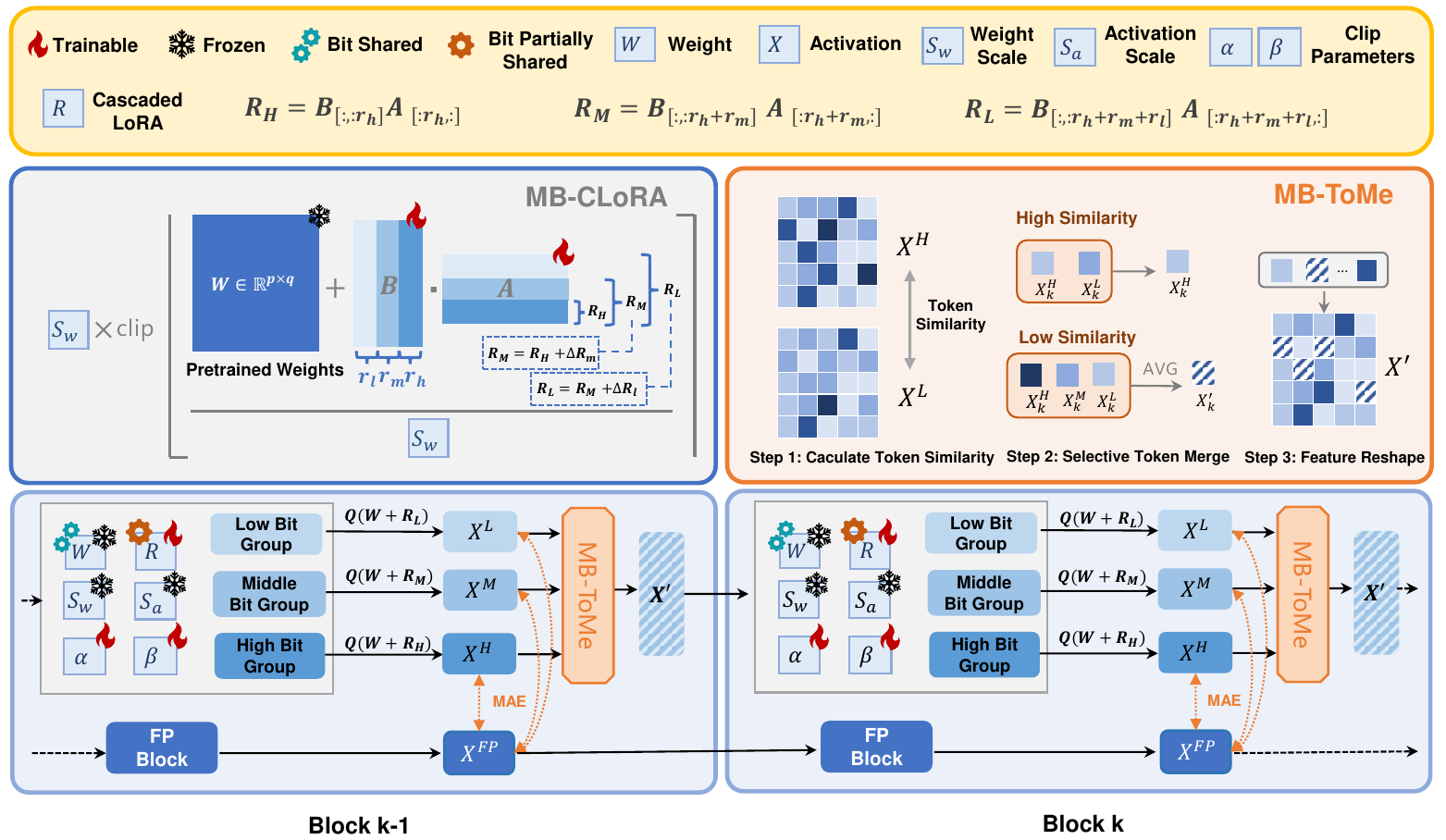}
\caption{Overview of our proposed QuEPT. QuEPT calibrates the low-rank compensation matrix $\bm R$ and weight clipping parameters $\bm \alpha$ and $\bm \beta$ in block-wise reconstruction, weights $\bm W $ and quantization scale $\bm S$ are frozen. The reconstruction process consists of two stages: (1) merging multiple bit-width features from different bit groups by Multi-Bit Token Merging (MB-ToMe); (2) optimizing multi-bit quantization loss using a Multi-Bit Cascaded Low-Rank Adapters (MB-CLoRA).}
\label{framework}
\end{figure*}

To address the above challenges, we propose a novel post-training elastic quantization framework, QuEPT, as shown in Figure~\ref{framework}. QuEPT introduces Multi-Bit Token Merging (MB-ToMe), which addresses competitive conflicts between bit-widths by selectively preserving robust high-precision token features while fusing less stable representations across precision levels to maintain balanced performance during multi-bit optimization. Complementarily, we propose Multi-Bit Cascaded Low-Rank Adapters (MB-CLoRA), redesigning the parameter-update pathway into a cascaded low-rank flow that lets gradients harvested at high precision gently diffuse into their low-precision counterparts, turning conflict into synergy. Our method provides a unified multi-bit width quantization solution for various transformer models, including ViTs, LLMs, and Multi-modal LLM (MLLM) architectures such as LLaVA~\cite{llava1,llavaonevision}. This approach only requires one-shot calibration and can be deployed in real time to any target bit-width by selecting the appropriate LoRA, completely eliminating re-optimization. Moreover, it can be seamlessly converted to a training-free mixed-precision scheme by utilizing layer sensitivity data to assign different bit widths across layers.   

Our contributions are summarized as follows:
\begin{itemize}
\item  We present QuEPT, an efficient elastic quantization framework for transformers. The elastic quantized model supports real-time configuration of different bit widths without the need to repeatedly optimize parameters.
\item We propose the Multi-Bit Token Merging (MB-ToMe) to reconstruct the mixed quantization error in each block by fusing tokens of different precision levels, thereby alleviating the contention conflict between different bit-widths and improving quantization robustness.
\item We introduce the Multi-Bit Cascaded Low-Rank Adapters (MB-CLoRA) to enhance the correlation between different bit-width groups, thereby improving the overall quantization performance of QuEPT.
\item Extensive experiments on ViT, LLM, and MLLM models verify that the performance of QuEPT is comparable to or even better than the current PTQ methods.
\end{itemize}

%% file: sections/2_related_works.tex
\section{Related Works}

\paragraph{Multi-Bit Quantization.}
Early multi-bit quantization are typically based on Quantization-Aware Training (QAT). For instance, RobustQuant~\cite{robustquant} discovered that uniformly distributed weights are more robust to varying bit-widths and introduced kurtosis regularization to encourage a more uniform distribution during training; Any-Precision~\cite{any-precision} flexibly converts a high-precision parent model into lower-bit-width models through truncation; EQ-Net~\cite{eqnet} employs both skewness and kurtosis regularization to constrain the weight distribution, combining them with a group guidance distillation loss to improve robustness across different bit widths. However, the training overhead of these QAT-based methods is prohibitive, making them unsuitable for modern large-scale Transformer models. Recently, several PTQ-based multi-bit quantization methods have emerged. Among them, PTMQ~\cite{ptmq} absorbs multi-bit errors through rounding and validates its effectiveness on CNNs and ViTs by combining feature mixing with group-wise distillation. Despite being a PTQ method, its quantization overhead remains substantial for LLMs and ViTs. Another approach, Any-Precision LLM~\cite{any-precision-llm}, proposed a Non-uniform Quantization-based incremental upscaling to adapt the Any-Precision method for LLMs; however, its performance at low bit-widths is limited, and it fails to account for activation quantization. MatQuant~\cite{matquant} introduced a training approach using co-distillation regularization, but it is suboptimal as it does not consider the competitive relationship between different bit widths. Our method improves robustness across different bit widths by integrating multi-bit token merging method and a cascaded LoRA strategy. This approach is designed to account for the importance of each bit width.

\paragraph{LoRA-based Quantization.}
The Low-Rank Adaptation (LoRA)~\cite{lora} method reduces the overhead of model fine-tuning by training two low-rank matrices that are added to the frozen pre-trained weights. Recent approaches~\cite{qlora,qalora,lqlora,loftq,owq} combine LoRA with quantized models, which compensate for quantization errors and achieve the effect of fine-tuning on the downstream dataset. LR-QAT~\cite{lrqat} adds LoRA to the pre-trained weights directly within the clipping operator, allowing the trained LoRA adapters to fuse back into the main weights after training. Furthermore, LoRA can also be used with PTQ. CBQ~\cite{cbq} replaces the weight-rounding matrix in AdaRound~\cite{adaround} with LoRA, reducing the overhead of the AdaRound approach in LLM. SVDQuant~\cite{svdquant} ingeniously combines SmoothQuant~\cite{smoothquant} and LoRA, tasking LoRA with learning an outlier branch to make the remaining weights easier to quantize. Our method is a PTQ approach that places the LoRA module inside the clipping operator. This ensures that the LoRA branch incurs no additional overhead during inference. 
Furthermore, we employ a cascaded LoRA training method for LoRA that are robust across all bit widths.

%% file: sections/3_method.tex
\section{Method}


\subsection{Post-Training Elastic Quantization Modeling}
We formulate QuEPT as a block-wise optimization task for post-training quantization. Specifically, given the target set of quantization bit-widths $\mathcal{B}=\{b | b \in \mathbb{Z^+}, m \le b \le n\}, |\mathcal{B}|\ge 3$, the optimization objective of Elastic Quantization can be expressed by the following formula:
\begin{equation}
\min_{\bm A, \bm B,\bm \alpha, \bm \beta} \sum_{b \in \mathcal{B}}\mathbb{E} [(F^l(\bm W; \bm X)-F_b^l(\bm {\hat W}_b|\bm{A}_b ,\bm{B}_b,\bm {\alpha}_b,\bm{\beta}_b;\bm {\hat X})].
\end{equation}
Here, $\bm A$ and $\bm B$ are the LoRA parameters, $\bm \alpha$ and $\bm \beta$ are the weight clipping parameters, which include all target bit widths. $F^l$ and $F^l_b$ denote the output of the $l$-th block of the full-precision model and the output of the $b$-bit quantized model, respectively. 
The complete Low-Rank-based quantization includes weight and activation quantization, which can be represented as follows:
\begin{equation}
 \bm {\hat W}_b = \bm{s_w}^{b} \cdot (clip(\lfloor \frac {\bm {W +BA}} {\bm{s_w}^{b}}\rceil +\bm {z_w}^{b}, -2^{b-1},2^{b-1}-1 ) -\bm{z_w}^{b} ),
\end{equation}
\begin{equation}
\bm {\hat X}_b = \bm{s_a}^{b} \cdot clip(\lfloor \frac {\bm X} {\bm{s_a}^{b} }\rceil, -2^{b-1},2^{b-1}-1 ),
\end{equation}
where $b$ represents a quantization bit width in the set $\mathcal{B}$, $\bm{s_w}$, $\bm{s_a}$ represents the weight and activation quantization scales respectively, and $\bm{z_w}$ denotes the corresponding weight zero-point. $clip()$ refers to a truncated operation. The operator $\left \lfloor \cdot \right \rceil$ rounds the continuous numbers to the nearest integers. Unlike QLoRA, we place the low-rank parameters directly into the quantization formula. Compared to rounding-based methods, the low-rank parameters are much smaller than the full parameter size and do not need to be restricted to the 0-1 range.
However, during the LoRA parameter update process, the quantized weight $\bm {\hat W}$ undergoes continuous changes. To accommodate these dynamic adjustments, we propose a method that optimizes the weight clipping threshold concurrently with the low-rank parameter. The formula  of the $b$-bit weight quantization scale is expressed as follows:
\begin{equation}
    \bm s_w^{b} =\frac {\bm \alpha_b \times max(\bm{W}+\bm{B}_b\bm{A}_b) - \bm \beta_b \times min(\bm{W}+\bm{B}_b\bm{A}_b)}{2^{b-1}},
\end{equation}

\begin{equation}
    \bm z_w^{b}= - \lfloor \bm \beta_b \times \frac {min(\bm{W}+\bm{B}_b\bm{A}_b)}{\bm  s_w^{b}}\rceil.
\end{equation}
By concurrently training the clipping parameters, we enable the LoRA parameters to be optimized more effectively. The clipping mechanism handles the large-magnitude errors from weight outliers, creating a smoother and more manageable error signal for the low-rank adaptation to correct.

\subsection{The Pipeline of QuEPT}
As shown in Figure~\ref{framework}, QuEPT conducts a block-wise reconstruction that starts by initializing the weight-clipping parameters and cascaded LoRA adapters jointly with the frozen full-precision weights. Throughout the pipeline, we partition the target bit-width set $\mathcal{B}$  into three tiers: low-bit ($\mathcal{B}_L$), mid-bit ($\mathcal{B}_M$) and high-bit ($\mathcal{B}_H$). At each step, one bit-width is randomly sampled from each tier, exposing the current block to a diverse combination of quantization granularities. The input to the block is produced by the preceding Multi-Bit Token Merging, which fuses the token-wise features from already reconstructed blocks while selectively anchoring the most robust high-bit tokens and softly merging the rest.  The block’s forward pass is evaluated under the three sampled bit widths and the reconstruction loss, aggregated across all of them and measured with MAE, is back-propagated only into the corresponding slice of the cascaded LoRA adapters and the clipping thresholds.  This simultaneous exposure to low-, mid- and high-bit quantization errors forces the adapters to co-optimize across the entire precision range, turning potential inter-bit conflicts into synergy and allowing a single calibration run to yield a model that can be deployed at any predefined bit-width simply by selecting the matching LoRA slice, without any further optimization.

\subsection{Multi-Bit Token Merging (MB-ToMe)}
MB-ToMe aims to design a token fusion strategy that maintains balanced performance across different bit-widths during multi-bit optimization, without favoring any specific precision level. We explored two key design dimensions: the merge operation (selection vs. fusion) and the merge policy (uniform vs. selective). Figure 2 demonstrates the three primary cases that were examined:
\begin{figure}[!htb]
\centering
\includegraphics[width=1.0\linewidth]{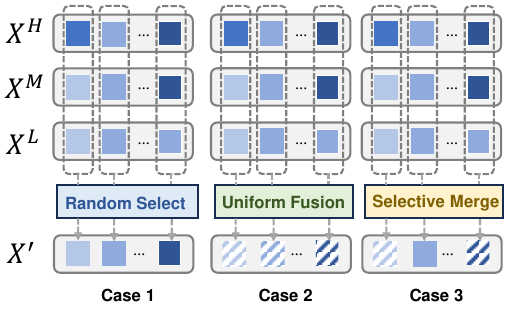}
\caption{Three cases of multi-bit token merging strategy.}
\label{token merging}
\end{figure}

\begin{itemize}
    \item[\textbullet] \textbf{Case 1: Random Selection.} For each token, we randomly select one feature representation from the entire set of available bit-widths. This represents a pure selection-based approach with a uniform policy, as every token and bit-width is treated with equal probability.

    \item[\textbullet] \textbf{Case 2: Uniform Fusion.} We explicitly divide the various bit width sets into three groups: low, medium, and high. After randomly selecting one bit-width from each group, we perform a weighted fusion of the corresponding feature outputs using a 1:1:1 ratio. This strategy shifts from discrete token selection to feature-level fusion, yet maintains a uniform fusion policy across all groups.

    \item[\textbullet] \textbf{Case 3: Selective Merge.} This strategy employs a selective mechanism to preserve high-precision information by identifying foundational tokens based on their quantization robustness, measured as the cosine similarity between their 8-bit and 4-bit representations. Tokens with high similarity are considered stable and are retained in their high-precision form, providing structural integrity to the feature map. This data-driven selection is more effective than random sampling. The remaining tokens are fused through a weighted average to maintain overall feature continuity. The hybrid approach thus preserves critical information via selective retention while ensuring smooth representation through fusion.

\end{itemize}

In multi-bit quantization, Case 1 with its random replacement strategy fails to produce well-integrated mixed features, leading to suboptimal representation quality. Case 2 mixes features from different bit-widths, retaining continuity but losing detailed information from high-bit-width tokens. Case 3 effectively combines both concepts, aiming to preserve the information from high-bit-width tokens as much as possible while also maintaining continuity between the different features. Table~\ref{tab:MB-ToMe} confirms that Case 3 is the most effective. This method can be formalized as follows:
\begin{equation}
\bm {X^{'}_k} =
\begin{cases}
\bm {X^H_k},& \text{if $k$ in set $\Phi$}  \\
\lambda_1 \bm {X^H_k} + \lambda_2 \bm {X^M_k} + \lambda_3 \bm {X^L_k}, & \text{else}
\end{cases}.
\end{equation}
Here, $\Phi$ is a set of indices. We first pre-calculate the token-wise cosine similarity between the features of the high-bit group $X_k^H$, and those of the low-bit group, $X_k^L$. By sorting the similarity scores for all tokens, we then save the indices corresponding to the top $p$\% of similarities as $\Phi$. $\{\lambda_1,\lambda_2,\lambda_3\}$ are hyperparameters that control the fusion ratio. Finally, all the $X_k^{'}$ are re-concatenated to form a feature map of the same size as the original one, which then serves as the quantization input for the next block. Figure 3 shows that merging dissimilar tokens leads to a more uniform numerical distribution that is more robust to varying bit-widths.

\begin{figure}[!htb]
\centering
\includegraphics[width=1.0\linewidth]{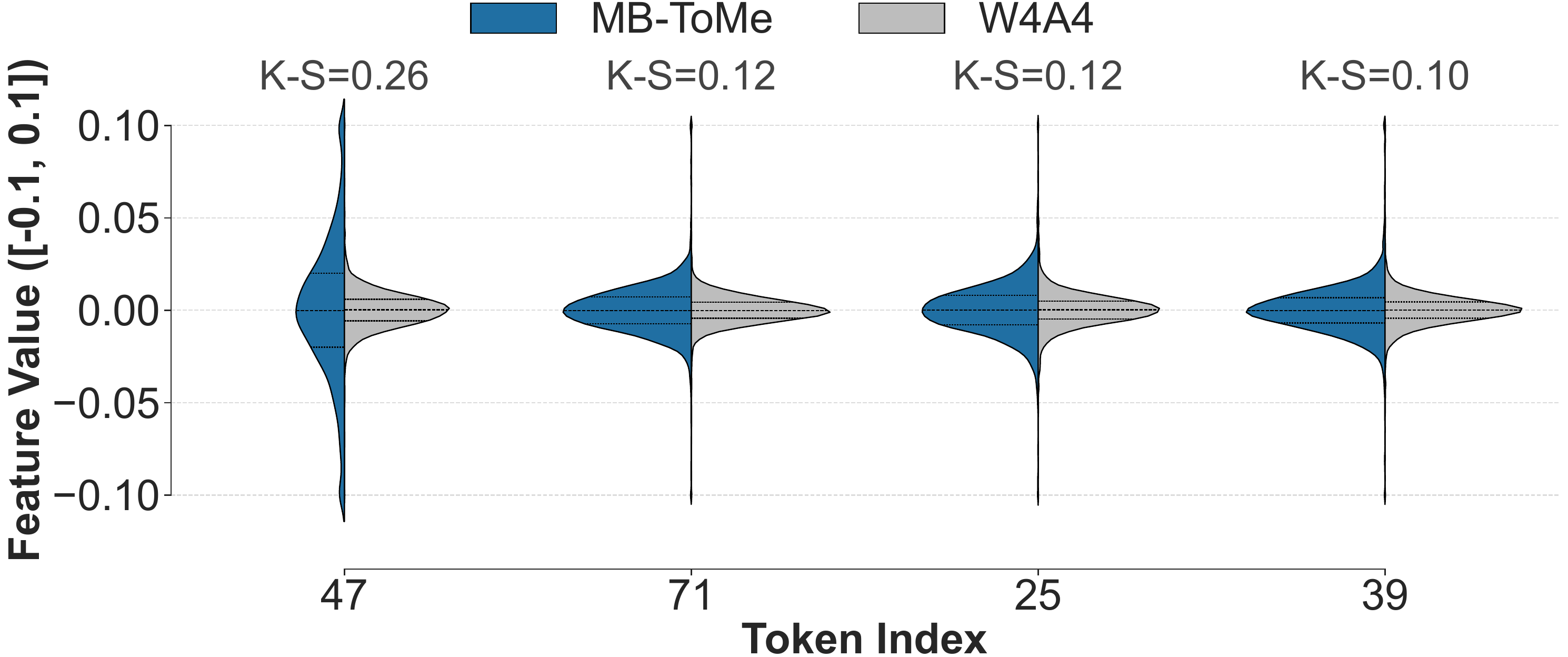}
\caption{Feature distribution comparison for top 4 divergent tokens on the input of block 1 of LLaMA2-7b. Token sorted by Kolmogorov-Smirnov (K-S) statistic.}
\label{token diff}
\end{figure}


\subsection{Multi-Bit Cascaded LoRA (MB-CLoRA)}

Cascaded low-rank adaptation structure enables efficient and hierarchical parameter sharing across multiple bit-widths. 
Rather than maintaining separate LoRA parameters for each bit-width, we construct a unified parameter space where the LoRA matrices $\bm{A} \in \mathbb{R}^{r \times q}$ and $\bm{B} \in \mathbb{R}^{p \times r}$   serve all bit-widths through a hierarchical structure. The key insight is that lower bit-widths, which suffer greater quantization error, can leverage the adaptation capacity optimized for higher bit-widths while adding specialized compensation.
Formally, for any given bit-width $b \in \mathcal{B}$, the corresponding low-rank compensation matrix $\bm{R}^{(b)}$ is defined as:
\begin{equation}
\bm{R}^{(b)} = \bm{B}_{[:,:r_b]} \bm{A}_{[:r_b,:]},
\end{equation}
where $r_b$ represents the effective rank for bit-width $b$ . Crucially, the rank allocation follows a cascaded pattern where $r_b$ increases monotonically as bit-width group decreases:
\begin{equation}
r_b = 
\begin{cases} 
    r_h & \text{if } b \in \mathcal{B}_H \\
    r_h + r_m & \text{if } b \in \mathcal{B}_M \\
    r_h + r_m + r_l & \text{if } b \in \mathcal{B}_L 
\end{cases} .
\end{equation}
This structure establishes a natural inheritance mechanism: when processing low-bit-width $b \in \mathcal{B}_L$, the compensation matrix $\bm{R}_L = \bm{B}_{[:,:r_h+r_m+r_l]} \bm{A}_{[:r_h+r_m+r_l,:]}$ will contain all parameters of medium and high bitwidth, because $\bm{R}_H = \bm{B}_{[:,:r_h]} \bm{A}_{[:r_h,:]}$ and $\bm{R}_M = \bm{B}_{[:,:r_h+r_m]} \bm{A}_{[:r_h+r_m,:]}$ will appear as leading submatrices.

During each training iteration, we simultaneously optimize across multiple bit-widths to ensure balanced adaptation. Specifically, we sample one bit-width from each tier: $b_H \sim \mathcal{B}_H$, $b_M \sim \mathcal{B}_M$, and $b_L \sim \mathcal{B}_L$. We then sequentially optimize the shared parameters $\bm{A}$ and $\bm{B}$ using each of these three bit-widths in turn. 
The unified optimization objective is to minimize the expected reconstruction error across all sampled bit-widths:
\begin{equation}
\min_{\bm{A},\bm{B},\bm{\alpha},\bm{\beta}} \sum_{b \in \{b_L,b_M, b_H \}} \left\| \bm{W}\bm{X} - \widehat{({\bm{W}} + \bm{R}^{(b)}) }{\bm{X}}^{'} \right\|_1 ,
\end{equation}
where $\widehat{({\bm{W}} + \bm{R}^{(b)})}$ denotes the joint quantization of the weight and its compensated LoRA, while $\bm X'$ are the merged features from the previous block, obtained via MB-ToMe. It is worth noting that we use Mean Absolute Error (MAE) loss instead of MSE loss, as our experiments (Table~\ref{tab:Ablation}) show that MAE generally yields superior results.




%% file: sections/4_experiments.tex
 \section{Experiments}
\input{tables/vit_main}

\subsubsection{Models and Evaluations.}
We evaluated our method on the Vision Transformer family such as ViT~\cite{vit}, DeiT~\cite{deit}, and Swin~\cite{swin}; the LLaMA~\cite{llama3} family of large language models; and the LLaVA-OneVision~\cite{llavaonevision} multimodal model. For the Vision Transformer models, we evaluated accuracy on ImageNet. For the LLaMA models, we assessed perplexity on the WikiText2~\cite{wikitext2} and C4~\cite{c4} datasets. Additionally, we evaluated accuracy on several zero-shot tasks, including the PIQA~\cite{piqa}, ARC-C~\cite{arc}, ARC-E~\cite{arc}, HellaSwag~\cite{hellaswag}, and WinoGrande~\cite{winogrande} datasets. For the multi-modal model evaluation, we selected the representative LLaVA-OV model and assessed its accuracy on the MMLU~\cite{mmmu}, TextVQA~\cite{textvqa}, VizWiz~\cite{vizwiz}, OCRBench~\cite{ocrbench}, and SEED~\cite{seed} datasets. %

\subsubsection{Calibration Datasets.}
For training the Vision Transformer models, we randomly selected 1024 unlabeled images from ImageNet as the calibration dataset; for the LLaMA family models, we used 128 samples from C4~\cite{c4} datasets for calibration; and for the LLaVA-OV model, we followed the same setup with MBQ~\cite{mbq}, also using 128 image-caption pairs sampled from the improved COCO Caption dataset proposed by ShareGPT4V~\cite{sharegpt4v} as the calibration dataset.

\subsubsection{Baselines.}
We compare with PTQ methods on Vision Transformers, such as PTQ4ViT~\cite{ptq4vit}, PDQuant~\cite{pdquant}, RepQ-ViT~\cite{repqvit},  ERQ~\cite{erq}, and PTMQ~\cite{ptmq}. For LLMs, we considered the quantization methods AWQ~\cite{awq}, GPTQ~\cite{gptq}, OmniQuant~\cite{omniquant}, AffineQuant~\cite{affinequant}, SmoothQuant~\cite{smoothquant}, QLLM~\cite{qllm}, QuaRot~\cite{quarot}, and DuQuant~\cite{duquant}. For multi-modal models, our primary comparison was with the MBQ~\cite{mbq} method.

\subsection{Main Results}
\subsubsection{Quantization of Vision Transformer Models.}
For weight-activation quantization on ViTs, we partition the bit-widths \{4,5,6,7,8\} into a low-bit group \{4\}, a mid-bit group \{5,6\}, and a high-bit group \{7,8\}. This same weight-activation quantization configuration is applied to other models, including LLMs and MLLMs. We use the reparameterization technique~\cite{repqvit} to initialize the activation quantizer. The quantization results for Vision Transformers under both single-bit and multi-bit scenarios are compared and summarized in Table~\ref{tab:vit-main}. 
For the ViT-S model, it achieves 6.2\% higher accuracy than ERQ on the W4A4 setting and a 1.46\% higher average accuracy across the five bit-widths. Unlike single-bit-width methods, which require five separate optimizations, our approach necessitates only one training run. As a result, the total training time is reduced to less than half of that required by ERQ. Our method is both more effective and significantly faster to train than the multi-bit quantization method PTMQ. For instance, on the ViT-B model, our method outperforms PTMQ by 6.1\%, 5.6\%, and 5.2\% on the W6A6, W7A7, and W8A8 settings, respectively. Compared to PTMQ, our method has significantly lower training overhead because it discards the computationally expensive rounding values, while requiring only 1/26 of the time overhead on a single Nvidia RTX 3090 GPU.


\subsubsection{Weight-Act Quantization of LLaMA Series Models.}

\input{tables/llama_wa_half}
Our method also demonstrates competitive performance in weight-activation quantization for the LLaMA family of models, as shown in Table~\ref{tab:llama wa half}. Since LLMs have outliers, we first apply a smoothing process with a Hadamard matrix~\cite{quarot} before quantization. On LLaMA2-7B with W4A4 quantization, our approach not only yields lower PPL on WikiText2 and C4 but also outperforms QuaRot and SpinQuant by 7.07\% and 0.28\%, respectively, in average accuracy across five benchmarks (PIQA, ARC-E, ARC-C, HellaSwag, and WinoGrande).

\subsubsection{Quantization of LLaVA-OneVision Model.}
\input{tables/llavaov}
Notably, for the weight-only quantization of LLaVA-OV and LLaMA, we explore a bit-width range of 2 to 8. The detailed settings are provided in the Appendix.
In the multimodal model LLaVA-OneVision-7B, QuEPT achieves better results than the current SOTA method MBQ, as shown in Table~\ref{tab:llavaov}. In a comparison of average accuracy across five datasets (MMMU, OCRBench, TextVQA, VizWiz, and SEED), QuEPT is 1.5\% higher than MBQ under W3A16 quantization and 4.3\% higher under W4A8 quantization.

\subsubsection{Results of Mixed Precision of LLaMA2-7B Model.}
\begin{figure}[!htb]
\centering
\includegraphics[width=1.0\linewidth]{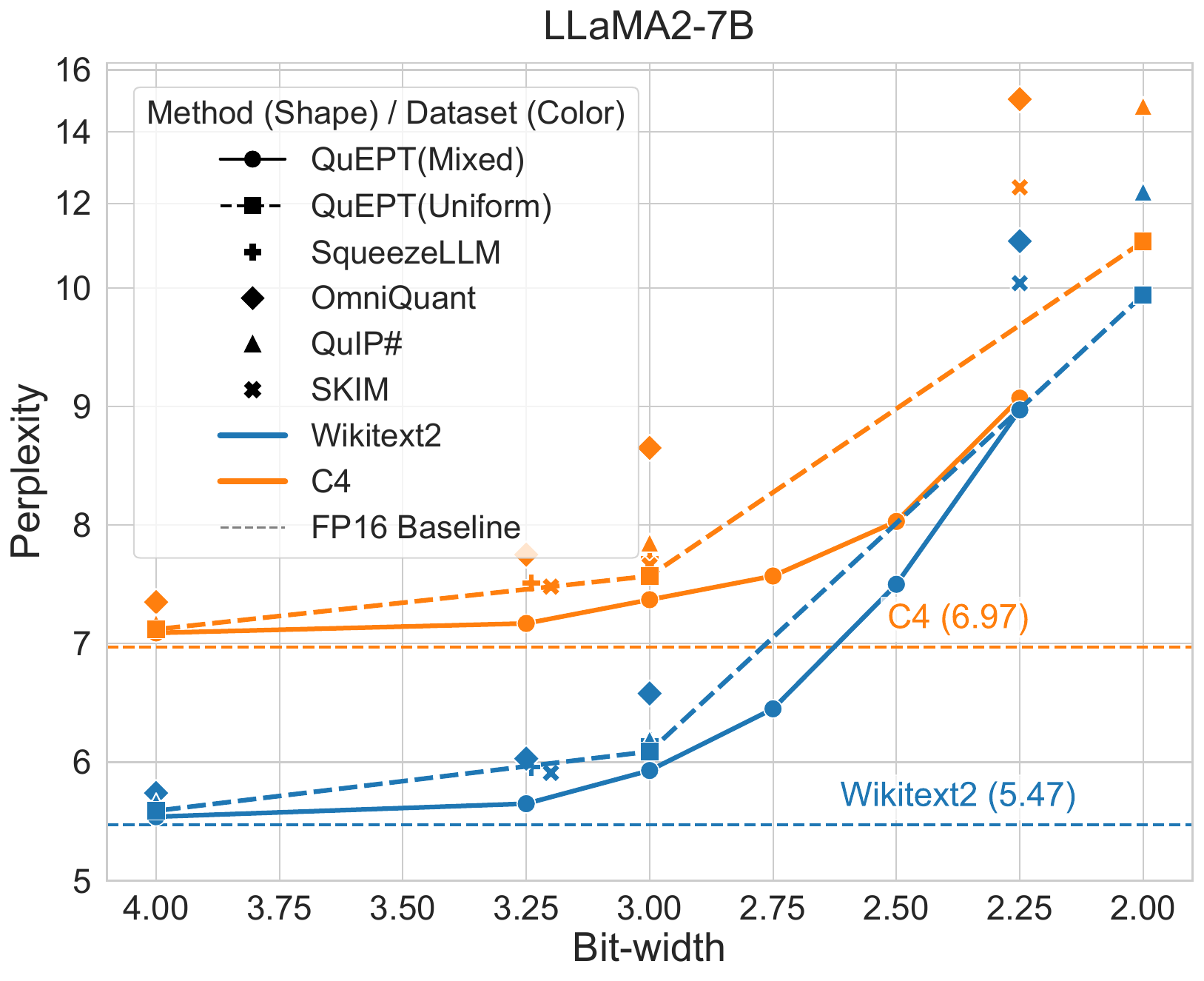}
\caption{Results of Mixed Precision of LLaMA2-7B.}
\label{mixed}
\end{figure}

As our method is an elastic quantization approach, it can be seamlessly transformed into a mixed-precision quantization method by simply modifying the LoRA configuration for each layer. To achieve this, we first use KL divergence to evaluate the per-layer sensitivity of the model. We then employ a DP algorithm to determine the optimal per-layer bit-width allocation that minimizes the total sensitivity, with our bit-widths ranging from 2 to 8 bits.
As shown in Figure~\ref{mixed}, we compare the PPL of our method on the LLaMA2-7B model against several weight mixed-precision and non-uniform quantization methods, including SKIM~\cite{skim}, SqueezeLLM~\cite{squeezellm}, and QuIP\#~\cite{quip}. The results indicate that with average bit-widths of 2.25/3.00/4.00, our method achieves a perplexity of only 8.97/5.93/5.54, respectively, on WikiText2. 

\subsection{Ablation Studies}

\subsubsection{The Effectiveness of Multi-Bit Cascaded LoRA.}
\input{tables/lora_share}
To validate the Effectiveness of MB-CLoRA, we designed two sets of experiments. We evaluated the accuracy of the ViT-S model on ImageNet and the average accuracy of the LLaMA2-7B model across five benchmarks. We compared three LoRA sharing strategies: (1) Case 1 (Fully Shared): All bit-widths share a single LoRA module. The LoRA rank was set to 12 for ViT-S and 48 for LLaMA2-7B. (2) Case 2 (Independent): The LoRA is partitioned by bit-width into a low-bit group \{4\}, a mid-bit group \{5, 6\}, and a high-bit group \{7, 8\}. For the ViT-S model, the rank for each group was 4, and for the LLaMA2-7B model, it was 16. (3) Case 3 (MB-CLoRA): This approach maintains the same total LoRA rank as the other settings but employs the cascaded sharing strategy described in the methods section.
The results in Table~\ref{tab:lora share} demonstrate that our proposed MB-CLoRA performs better at lower bit-widths. For instance, on LLaMA2-7B with W4A4, MB-CLoRA's accuracy is 2.4\% and 0.7\% higher than that of the Independent and Fully Shared strategies, respectively. At higher bit-widths, the performance of MB-CLoRA is comparable to the Independent strategy.

\subsubsection{The Effectiveness of Multi-Bit Token Merging.}
\input{tables/MB-ToMe}
We compared the performance of several multi-bit-width feature fusion strategies. The results in Table~\ref{tab:MB-ToMe} confirm that Case 3 achieves the best performance across all bit-widths. Compared to Case 1, Case 3 prevents random feature selection from affecting the results. In contrast to Case 2, Case 3 preserves high-bit features as much as possible, leading to more accurate feature information. 
Therefore, Case 3 yields the best results. For instance, on the LLaMA2-7B model, MB-ToMe shows performance gains of 5.9\%, 0.8\%, and 0.2\% over the Case 1 method at 4, 6, and 8 bits, respectively.

\subsubsection{The Effectiveness of Different Modules.}
As presented in Table~\ref{tab:Ablation}, we evaluated combinations of different modules. Our baseline uses 3 independent LoRAs ($r$=4) with MSE loss to compensate for low, medium, and high bit-width groups. With MB-CLoRA, W4A4 accuracy increased, while W6A6 and W8A8 accuracy dropped. This is because MB-CLoRA boosts the effective rank for low bit-widths via rank sharing, but this optimization harms the LoRA compensation for medium and high bit-widths. The combination of MB-CLoRA and MB-ToMe led to an improvement in low-bit-width performance.
\input{tables/ablation}

%% file: tables/vit_main.tex
\begin{table*}[!htb]
  \centering
  \small
\begin{tabular}{cccccccccccc}
    \toprule
    \multirow{2}[4]{*}{\textbf{Model}} & \multirow{2}[4]{*}{\textbf{Method}} & \multirow{2}[4]{*}{\textbf{Opti.}} & \multirow{2}[4]{*}{\textbf{Uniform}} & \multirow{2}[4]{*}{\textbf{Criterion}} & \multicolumn{5}{c}{\textbf{Bit-Width}} & \multirow{2}[4]{*}{\textbf{Time (min)}} & \multirow{2}[4]{*}{\textbf{FP32}} \\
\cmidrule{6-10}          &       &       &       &       & W4A4  & W5A5  & W6A6  & W7A7  & W8A8  &       &  \\
    \midrule
    \multirow{6}[4]{*}{ViT-S} & PTQ4ViT & \ding{55} & \ding{55} & \multirow{4}[2]{*}{Single-Bit} & 42.6  & 72.7  & 78.6  & 80.5  & 81.0  & 7×N & \multirow{6}[4]{*}{81.3} \\
          & PD-Quant & \ding{51} & \ding{51} &       & 21.2  & 65.1  & 71.4  & --     & --     & 110×N &  \\
          & RepQ-ViT & \ding{55} & \ding{55} &       & 65.1  & 78.4  & 80.4  & 80.8  & 81.2  & 4×N &  \\
          & ERQ   & \ding{51} & \ding{55} &       & 68.9  & 78.8  & 80.5  & 80.9  & 81.2  & 9×N &  \\
\cmidrule{2-11}          & PTMQ  & \ding{51} & \ding{51} & \multirow{2}[2]{*}{Multi-Bit} & --     & --     & 76.1  & 77.1  & 78.2  & 430 &  \\
          & \textbf{QuEPT}   & \ding{51} & \ding{51} &       & \textbf{75.1} & \textbf{79.7} & \textbf{80.6} & \textbf{81.0} & \textbf{81.2} & 17 &  \\
    \midrule
    \multirow{6}[4]{*}{ViT-B} & PTQ4ViT & \ding{55} & \ding{55} & \multirow{4}[2]{*}{Single-Bit} & 30.7  & 72.3  & 81.7  & 83.7  & 84.3  & 13×N & \multirow{6}[4]{*}{84.5} \\
          & PD-Quant & \ding{51} & \ding{51} &       & 34.9  & 58.4  & 63.1  & --     & --     & 230×N &  \\
          & RepQ-ViT & \ding{55} & \ding{55} &       & 68.5  & 82.0  & 83.6  & 84.3  & 84.4  & 8×N &  \\
          & ERQ   & \ding{51} & \ding{55} &       & 76.6  & 82.8  & \textbf{83.9}  & \textbf{84.3}  & \textbf{84.4}  & 17×N &  \\
\cmidrule{2-11}          & PTMQ  & \ding{51} & \ding{51} & \multirow{2}[2]{*}{Multi-Bit} & --     &  -- & 77.7  & 78.6  & 79.1  & 950 &  \\
          & \textbf{QuEPT}   & \ding{51} & \ding{51} &       & \textbf{80.7} & \textbf{83.3} & 83.8 & 84.2 & 84.3 & 36 &  \\
    \midrule
    \multirow{6}[4]{*}{DeiT-S} & PTQ4ViT & \ding{55} & \ding{55} & \multirow{4}[2]{*}{Single-Bit} & 34.1  & 70.3  & 76.3  & 79.0  & 79.5  & 7×N & \multirow{6}[4]{*}{79.8} \\
          & PD-Quant & \ding{51} & \ding{51} &       & 64.9  & 74.9  & 77.6  & --     & --     & 110×N &  \\
          & RepQ-ViT & \ding{55} & \ding{55} &       & 69.0  & 77.0  & 78.9  & 79.5  & 79.7  & 4×N &  \\
          & ERQ   & \ding{51} & \ding{55} &       & 72.6  & 77.6  & 79.0  & 79.5  & 79.6  & 9×N &  \\
\cmidrule{2-11}          & PTMQ  & \ding{51} & \ding{51} & \multirow{2}[2]{*}{Multi-Bit} & --     & --     & 78.7  & 79.3  & 79.5  & 430 &  \\
          & \textbf{QuEPT}  & \ding{51} & \ding{51} &       & \textbf{75.3} & \textbf{78.1} & \textbf{79.3} & \textbf{79.7} & \textbf{79.7} & 17 &  \\
    \midrule
    \multirow{5}[4]{*}{Swin-S} & PTQ4ViT & \ding{55} & \ding{55} & \multirow{4}[2]{*}{Single-Bit} & 76.1  & 80.9  & 82.4  & 82.9  & 83.1  & 17×N & \multirow{5}[4]{*}{83.2 } \\
          & PD-Quant & \ding{51} & \ding{51} &       & 70.4  & 81.3  & 82.3  & --     & --     & 270×N &  \\
          & RepQ-ViT & \ding{55} & \ding{55} &       & 79.5  & 82.1  & 82.8  & 83.0  & 83.0  & 8×N &  \\
          & ERQ   & \ding{51} & \ding{55} &       & 80.7  & 82.4  & 82.9  & 83.1  & 83.2  & 20×N &  \\
\cmidrule{2-11}          & \textbf{QuEPT}  & \ding{51} & \ding{51} & Multi-Bit & \textbf{81.9} & \textbf{82.8} & \textbf{82.9} & \textbf{83.1} & \textbf{83.2} & 38 &  \\
    \bottomrule
    \end{tabular}%
    \caption{Comparison of the top-1 accuracy across various ViTs on ImageNet.``Opti." indicates whether a method is optimization-based, and ``Uniform" indicates whether it uses a uniform quantizer. The best results are highlighted in bold.}
    \label{tab:vit-main}%
    
\end{table*}%

%% file: tables/llama_wa_half.tex
\begin{table}[!htb]
  \centering
  \resizebox{\columnwidth}{!}{
    \begin{tabular}{cccccc}
    \toprule
    \multirow{3}[2]{*}{\textbf{Model}} & \multirow{3}[2]{*}{\textbf{Bits}} & \multirow{3}[2]{*}{\textbf{Method}} & \multirow{3}[2]{*}{\textbf{Wiki.(↓)}} & \multirow{3}[2]{*}{\textbf{C4($\downarrow$)}} & \multirow{3}[2]{*}{\textbf{0-shot$^5$ Avg.($\uparrow$)}} \\
          &       &       &       &       &  \\
          &       &       &       &       &  \\
    \midrule
    \multirow{11}[10]{*}{L2-7B} & FP16  & -     & 5.47  & 6.97  & 65.71  \\
\cmidrule{2-6}          & \multirow{5}[2]{*}{W4A4} & OmniQ & 14.26  & 18.02  & 49.13  \\
          &       & QLLM  & 11.75  & 13.26  & 51.60  \\
          &       & QuaRot & 9.66  & 11.98  & 54.55  \\
          &       & DuQuant & \textbf{6.28}  & 7.90     & 59.11  \\
          &       & \textbf{QuEPT}   & 6.33  & \textbf{7.86}  & \textbf{61.62}  \\
\cmidrule{2-6}          & W5A5  & \textbf{QuEPT}   & 5.66  & 7.17  & 64.42  \\
\cmidrule{2-6}          & \multirow{3}[2]{*}{W6A6} & SmoothQ & 6.20  & 7.76  & 61.40  \\
          &       & OmniQ & 5.87  & 7.48  & 58.56  \\
          &       & \textbf{QuEPT}   & \textbf{5.53}  & \textbf{7.03}  & \textbf{65.48}  \\
\cmidrule{2-6}          & W8A8  & \textbf{QuEPT}   & 5.48  & 6.98  & 66.23  \\
    \midrule
    \multirow{10}[10]{*}{L2-13B} & FP16  & -     & 4.88  & 6.46  & 69.35  \\
\cmidrule{2-6}          & \multirow{4}[2]{*}{W4A4} & OmniQ & 12.30  & 14.55  & 53.13  \\
          &       & QLLM  & 9.09  & 11.13  & 54.31  \\
          &       & DuQuant & \textbf{5.42}  & \textbf{7.05}  & 62.84  \\
          &       & \textbf{QuEPT}   & 5.53  & 7.16  & \textbf{66.05}  \\
\cmidrule{2-6}          & W5A5  & \textbf{QuEPT}   & 5.10  & 6.69  & 68.39  \\
\cmidrule{2-6}          & \multirow{3}[2]{*}{W6A6} & SmoothQ & 5.18  & 6.76  & 64.30  \\
          &       & OmniQ & 5.14  & 6.74  & 64.71  \\
          &       & \textbf{QuEPT}   & \textbf{5.00}  & \textbf{6.59}  & \textbf{68.72}  \\
\cmidrule{2-6}          & W8A8  & \textbf{QuEPT}   & 4.94  & 6.53  & 69.09  \\
    \midrule
    \multirow{12}[10]{*}{L3-8B} & FP16  & -     & 6.14  & 8.88  & 72.65  \\
\cmidrule{2-6}          & \multirow{5}[2]{*}{W4A4} & OmniQ & 3.64e3 & 2.80e3 & 35.70  \\
          &       & AffineQ & 2.12e4 & 3.46e4 & 35.44  \\
          &       & QuaRot & 10.41  & 14.33  & 57.16  \\
          &       & DuQuant & 8.56  & 11.98     & 65.05  \\
          &       & \textbf{QuEPT}   & \textbf{8.25}  & \textbf{11.67}  & \textbf{67.04}  \\
\cmidrule{2-6}          & W5A5  & \textbf{QuEPT}   & 6.76  & 9.72  & 70.99  \\
\cmidrule{2-6}          & \multirow{4}[2]{*}{W6A6} & SmoothQ & 7.07  & 9.57  & 70.50  \\
          &       & OmniQ & 7.24  & 9.82  & 69.51  \\
          &       & AffineQ & 7.35  & 9.99  & 69.22  \\
          &       & \textbf{QuEPT}   & \textbf{6.34}  & \textbf{9.12}  & \textbf{71.56}  \\
\cmidrule{2-6}          & W8A8  & \textbf{QuEPT}   & 6.20  & 8.96  & 72.87  \\
    \bottomrule
    \end{tabular}%
    
    }
    \caption{Weight-Activation quantization results of LLaMA models. We report the perplexity on WikiText2 and C4 (↓ indicates lower is better), and the average accuracy (Avg.) denotes the average accuracy across 5 datasets.}
  \label{tab:llama wa half}%
\end{table}%

%% file: tables/llavaov.tex
\begin{table}[!htb]
  \centering
  \small
  \resizebox{\columnwidth}{!}{
    \begin{tabular}{ccccccc}
    \toprule
    \multirow{2}[2]{*}{\textbf{Bits}} & \multirow{2}[2]{*}{\textbf{Method}} & \multirow{2}[2]{*}{\textbf{mmmu}} & \multirow{2}[2]{*}{\textbf{ocrbench}} & \multirow{2}[2]{*}{\textbf{textvqa}} & \multirow{2}[2]{*}{\textbf{vizwiz}} & \multirow{2}[2]{*}{\textbf{seed}} \\
          &       &       &       &       &       &  \\
    \midrule
    FP16  & -     & 48.0  & 62.2  & 76.1  & 60.4  & 72.6  \\
    \midrule
    W2 & \textbf{QuEPT}   & 32.2  & 47.8  & 64.0  & 55.2  & 67.0  \\
    \midrule
    \multirow{3}[2]{*}{W3} & AWQ   & 36.6  & 59.3  & 70.0  & 56.4  & 51.5  \\
          & MBQ   & 42.0  & \textbf{61.1} & 73.3  & \textbf{60.7} & 66.4  \\
          & \textbf{QuEPT}   & \textbf{44.6} & 60.6  & \textbf{74.1} & 60.3  & \textbf{71.6} \\
    \midrule
    W4A4  & \textbf{QuEPT}   & 33.9  & 49.1  & 64.4  & 59.8  & 66.5  \\
    \midrule
    \multirow{3}[2]{*}{W4A8} & SmoothQ & 30.9  & 32.0  & 56.9  & 56.7  & 41.6  \\
          & MBQ   & 42.6  & 52.3  & 68.3  & 58.9  & 64.4  \\
          & \textbf{QuEPT}   & \textbf{43.4} & \textbf{61.2} & \textbf{71.5} & \textbf{61.3} & \textbf{70.7} \\
    \bottomrule
    \end{tabular}%
    }
    \caption{Weight-only and weight-activation quantization results of LLaVA-OneVision-7B model. }
    \label{tab:llavaov}%
\end{table}%

%% file: tables/lora_share.tex
\begin{table}[!htb]
  \centering
  \resizebox{\columnwidth}{!}{
  \small
    \begin{tabular}{ccccccc}
    \toprule
    \textbf{Model} & \textbf{Method} & \textbf{W4A4} & \textbf{W5A5} & \textbf{W6A6} & \textbf{W7A7} & \textbf{W8A8} \\
    \midrule
    \multirow{3}[2]{*}{ViT-S} & Case 1 & 73.7  & 78.6  & 80.6  & 79.9  & 81.1  \\
          & Case 2 & 73.5  & 78.9  & \textbf{80.7} & \textbf{81.1} & \textbf{81.2} \\
          & \textbf{Case 3} & \textbf{74.7} & \textbf{79.4} & 80.6  & 81.0  & 81.1  \\
    \midrule
    \multirow{3}[2]{*}{L2-7B} & Case 1 & 60.9  & 64.2  & 64.9  & 65.2  & 65.5 \\
          & Case 2 & 59.2  & 64.4  & 65.3  & \textbf{65.7} & \textbf{65.7} \\
          & \textbf{Case 3} & \textbf{61.6} & \textbf{64.5} & \textbf{65.5} & 65.6  & 65.6 \\
    \bottomrule
    \end{tabular}%
    }
     \caption{Ablation studies for Multi-Bit Cascaded LoRA.}
  \label{tab:lora share}%
\end{table}%

%% file: tables/MB-ToMe.tex
\begin{table}[!htb]
  \centering
  \resizebox{\columnwidth}{!}{%
    \begin{tabular}{clccccc}
    \toprule
    \textbf{Model} & \multicolumn{1}{c}{\textbf{Method}} & \textbf{W4A4} & \textbf{W5A5} & \textbf{W6A6} & \textbf{W7A7} & \textbf{W8A8} \\
    \midrule
    \multirow{3}[2]{*}{ViT-S} & Case 1 & 74.5  & 79.4  & 80.3  & 80.5  & 80.9  \\
          & Case 2 & 74.9  & 79.3  & 80.2  & 80.6  & 80.8  \\
          & \textbf{Case 3} & \textbf{75.1} & \textbf{79.7} & \textbf{80.6} & \textbf{81.0} & \textbf{81.2} \\
    \midrule
    \multirow{3}[2]{*}{L2-7B} & Case 1 & 55.7  & 63.8  & 64.7  & 64.9  & 65.4  \\
          & Case 2 & 60.2  & 64.4  & 65.2  & 65.5  & 65.5  \\
          & \textbf{Case 3} & \textbf{61.6} & \textbf{64.5} & \textbf{65.5} & \textbf{65.6} & \textbf{65.6} \\
    \bottomrule
    \end{tabular}%
}
  \caption{Ablation studies for Multi-Bit Token Merging.}
  \label{tab:MB-ToMe}
\end{table}%

%% file: tables/ablation.tex
\begin{table}[!htb]
  \centering
  \resizebox{\columnwidth}{!}{
    \begin{tabular}{ccccccc}
    \toprule
    \multicolumn{1}{l}{Clip} & MB-CLoRA & MB-ToMe & MAE   & W4A4  & W6A6  & W8A8 \\
    \midrule
    \ding{55} & \ding{55} & \ding{55} & \ding{55} & 69.2  & 80.1  & 81.1  \\
    \ding{52} & \ding{55} & \ding{55} & \ding{55} & 70.1  & 80.2  & 81.1 
    \\
    \ding{52} & \ding{52} & \ding{55} & \ding{55} & 72.2  & 79.2  & 80.7  \\
    \ding{52} & \ding{55} & \ding{52} & \ding{55} & 73.0  & 80.1  & 80.9  \\
    \ding{52} & \ding{52} & \ding{52} & \ding{55} & 73.6  & 80.1  & 81.0  \\
    \ding{52} & \ding{52} & \ding{52} & \ding{52} & 74.7  & 80.7  & 81.1  \\
    \bottomrule
    \end{tabular}%
    }
    \caption{QuEPT modules ablation study on ViT-S model.}
  \label{tab:Ablation}%
\end{table}%

%% file: sections/5_conclusion.tex
\section{Conclusion}
This paper presents QuEPT, the first multi-bit-width PTQ method validated on ViTs, LLMs, and MLLMs. First, we introduce MB-CLoRA, which uses a cascaded LoRA to enable efficient parameter collaboration across bit-widths. Second, we propose MB-ToMe, a selective feature retention and fusion mechanism that safeguards performance from the negative impact of low-bit quantization. Finally, we have validated our method on various models, establishing it as a state-of-the-art multi-bit-width quantization approach. Nevertheless, our approach is not without its limitations. We have not explicitly handled outliers in LLMs, meaning that integrating our implementation with outlier mitigation techniques (e.g., SpinQuant~\cite{spinquant}) would likely yield better results. Additionally, our method's performance remains constrained at extremely low bit-widths, which requires further exploration in the future.